*Commentary*

# The Power of Modeling—a Response to PDDL2.1

**Fahiem Bacchus**                                       FBACCHUS@CS.TORONTO.EDU
*Department. of Computer Science*
*6 King's College Road*
*University Of Toronto*
*Toronto, Ontario*
*Canada, M5S 3H5.*

## Abstract

In this commentary I argue that although PDDL2.1 is a very useful standard for the planning competition, its design does not properly consider the issue of domain modeling. Hence, I would not advocate its use in specifying planning domains outside of the context of the planning competition. Rather, the field needs to explore different approaches and grapple more directly with the problem of effectively modeling and utilizing *all* of the diverse pieces of knowledge we typically have about planning domains.

## 1. Introduction

Fox and Long did a terrific job of organizing the 2002 Planning Competition. A non-trivial component of that work was specifying an extension of PDDL so that a much more interesting range of problems could be addressed in the competition. Although the extension represents a very useful standard for the competition, its design ignores both the power and reality of domain modeling. I will argue that some of the new features of PDDL2.1 are unnecessary: similar effects can just as easily be captured by more robust modeling of the planning domain.

My own TLPLAN planning system competed in the 2002 planning competition. Despite the fact that TLPLAN's specification language had no direct support for some features of PDDL2.1, we were still able to encode all of the competition domains. TLPLAN utilizes a language that is designed to be suitable for robustly modeling planning domains.[1] We found that many of the new features of PDDL2.1 used in the competition were in fact easy to capture simply by more robust modeling of the planning domain.

That PDDL2.1 takes the approach of adding new features to the language, rather than requiring that the domain be more robustly modeled, is perhaps not surprising given the dichotomy that still persists in the AI planning field. That dichotomy is between work on "domain-independent" planning and "control-intensive" planning. Just as work on control-intensive planning tends to ignore the applicability and power of state-of-the-art search algorithms for planning, work on domain-independent planning tends to ignore the power to be gained, and the requirements imposed, by domain modeling. Most planning researchers freely acknowledge the importance of both components, however, one rarely finds work at

---

1. TLPLAN also includes constructs for expressing domain specific control information, but I am not referring here to that component of the language.

```
(:durative-action load-truck
   :parameters (?t - truck)
               (?l - location)
               (?o - cargo)
               (?c - crane)
   :duration (= ?duration 5)
   :precondition (and (at ?t ?l)
                      (at ?o ?l)
                      (empty ?c))
   :effect       (and (loading ?t)
                      (at end (not loading ?t))
                      ...
)
(:action move-truck
   :parameters (?t - truck)
               ...
   :precondition (and (not (loading ?t))
                      ...)
)
```

Figure 1: Converting `(over all)` conditions

the interface of these two issues, nor are there many researchers who work on both of these issues.[2]

In the rest of this commentary I will elaborate on my argument by presenting examples of features and approaches appearing in PDDL2.1 that demonstrate its insensitivity to domain modeling.

## 2. Coordinating Concurrent Actions

In Section 5 of the PDDL2.1 document various constructs are presented to support durative actions. These are actions whose effects can either be immediate (like ordinary non-durative actions) or can occur at the end of some fixed duration. The duration can be either a constant or specified by some functional term. The obvious extension, used, e.g., by Bacchus & Ady (2001), where an action could have a range of effects each at a different future timepoint, was not included in PDDL2.1.[3] Without this extension the `(at start)` and `(at end)` specifiers are reasonable ways of specifying delayed effects.

With non-instantaneous actions concurrency control becomes an issue. To achieve this kind of control PDDL2.1 provides the `:condition` constructs `(at start)`, `(at end)`, and `(over all)`. TLPLAN does not provide the `(at end)` and `(over all)` constructs (`(at start)` is simply an ordinary precondition). Yet we were still able to model all of the planning domains without them. I would argue that in general these constructs are not necessary.

---

2. I am just as guilty of this as most of my colleagues.
3. Instead one must ascend to the much more complex specification of continuous effects (presented in Section 5.3 of the PDDL2.1 document) to achieve this degree of flexibility.

**2.1 `(over all)`:**

Consider the `load-truck` action given in Figure 6 of the pddl2.1 article. It contains the condition `(over all (at ?t ?l))`: i.e., the truck must stay at the same location through out the load. Why must it not be moved? Because we are not allowed to move a vehicle while it is being loaded. Similarly, we should not drive a car while it is being refueled, we should not turn off the oven while it is being used to bake a cake, we should not attempt to tie our shoelaces while running, etc. Our knowledge abounds with such common sense conditions. In natural language we use progressive tenses to refer to ongoing activities. Similarly, the most natural way to model an `(over all)` condition is simply to have the action assert that an activity is ongoing, and use the negation of that activity as a precondition for actions that could interfere. The resulting transformation of the `load-truck` action is given in Figure 1.

In Figure 1 the `load-truck` action advertises that it has initiated an on-going loading of the truck by adding a `loading` predicate to the state. At the end of the action this predicate is deleted. Actions like `move-truck` that could interfere with `loading` are blocked by adding `(not loading)` to their preconditions. Using this technique we were able to replace all `(over all)` conditions used in the competition domains. Interestingly the replacements made the domain more sensible and more readable.

Consider in brief the advantages of modeling `(over all)` effects by adding "progressive predicates" to the state.

1. The method can be realized without extending the standard strips/adl semantics.

2. An action's preconditions still encapsulates all of its interactions with the other actions. In the presence of `(over all)` conditions, one would have to examine every other action to see if any of its `(over all)` conditions are interfered with by one of the action's effects. Put another way, the action's activation condition continues to be Markovian, i.e., dependent only on the current state. The current state continues to carry all of the information needed to determine if an action can be applied; with `(over all)` conditions, one also needs to examine all currently active actions—the state no longer encapsulates all of the necessary information.

3. Finally, it seems to me that the resulting domain models are more natural and easier to understand.

### 2.1.1 Modularity

One concern with the above approach to avoiding `(over all)` conditions, is that it appears to make adding new actions to the domain non-modular.[4] Non-modularity potentially arises both from adding new actions that could be interfered with by previous actions, and from adding new actions that could interfere with previous actions.

The first case arises when we add new actions like `refuel`, `repair`, `change-tire`, etc., all of which can be interfered with by the existing action `move-truck`. Our solution of adding non-interference preconditions would seem to require modifying the description of

---

4. Thanks to David Smith and Martha Pollock for pointing out that I needed to address the issue of modularity.

`move-truck` to add `(not (refueling))`, `(not (repairing))`, `(not (changing-tire))`, etc., preconditions—a new precondition for every new action added.

The second case arises when we add a new action like `tow-truck` that also changes the location of the truck. We would then have to ensure that we add to its preconditions all progressives required to block it from interfering with previously defined actions. This second case is perhaps not as problematic, since it does not require modifying any of the old actions. However, the first case is an issue since one might not want to modify the definition of previous actions that had already been debugged.

When using `(over all)` conditions we need not make any changes to old actions nor worry about the effects of the new actions on the old. However, I would argue that this modularity exists only at the syntactic level—it is syntactically easy to modify the domain description to accommodate new actions. There is no corresponding modularity at the semantic level: the interferences between the new action and the old still exists. In most cases we cannot simply ignore these interactions, leaving it up to the planner and the `(over all)` conditions to resolve. As we have found when developing domains, in many cases when a new action is added to the domain a bug in the domain specification appears. For example, plans one would expect to find are no longer found by the planner. Often the bug lies in the new action, but just as often a bug is found in the specification of the old actions. If the domain no longer operates as expected, one is still left with the task of unraveling the interactions of the specification. In general, specifying a rich domain requires understanding the possible interactions in the domain, and in that task the `(over all)` conditions do not help.

It could be argued that it is the job of the planner to unravel the interactions in the domain. But this argument, I believe, trivializes the job of specifying a planning domain. The planner's job is to compute the interactions between actions in a sequence (or more complex composition) of actions. Getting the domain correctly specified is a difficult task, and requires at least understanding how actions interact *statically*, even if one can leave the dynamic interactions up to the planner.

Fortunately, in most systems the interactions between actions are relatively local: there is typically a relatively structured way in which actions can interfere with each other. This is what makes specifying planning domains feasible.

One can take advantage of this structure to build robust domain models that provide the advantages of semantic as well as syntactic modularity. A critical component of building good domain models is the ability to use definitions (axioms), a feature that is not provided by PDDL2.1. In TLPLAN, e.g., one can define a new predicate symbol using a first-order formula over previous defined symbols. By defining the right high level constructs one can typically provide a more explicit representation of the structure of interactions in the domain. The advantage of doing this is that one also obtains a declarative representation of this structure, thus achieving a more natural and easier to understand domain specification.

In the example above, one could define a new predicate `(must-be-stationary ?t)` to be the disjunction

```
(or (loading ?t) (changing-tire ?t) (repairing ?t) (refueling ?t)).
```

Now the `(move-truck ?t ?l1 ?l2)` action need only have a single precondition `(not (must-be-stationary ?t))`, and any new action that requires that the truck be station-

```
(:durative-action load-truck
   :parameters (?t - truck)
               (?l - location)
               (?o - cargo)
               (?c - crane)
   :duration (= ?duration 5)
   :precondition (and (at ?t ?l)
                      (at ?o ?l)
                      (empty ?c))
   :effect       (and (loading ?t)
                      (at end (not loading ?t))
                      (holding ?c ?o)
                      (not (at ?o ?l))
                      (at end (when (holding ?c ?o) (in ?o ?t)))
                      (at end (when (holding ?c ?o) (not (holding ?c ?o))))
                      (at end (when (not (holding ?c ?o)) (load-failed))))
)
```

Figure 2: Converting (at end) conditions

ary can be accommodated by simply adding a new disjunct to the definition of must-be-stationary. This approach has the advantage of explicitly introducing a new concept must-be-stationary, which helps in understanding and structuring the domain. In contrast, with (over all) conditions one only has the concept of "not changing at". In this simple example the difference appears to be trivial, but the key idea is that once we have a new concept like must-be-stationary in the domain, we can use it to build up more complicated concepts.

Using explicit progressive preconditions also allows for the coordination of far more complex shared uses of a resource. For example, we can specify that both refueling and driving require exclusive use of the truck, and that changing a tire can be done concurrently with repairing the truck but not concurrently with loading. These conditions can be accommodated by having each action explicitly mention the excluded activities in its precondition, or through axioms grouping and structuring these activities into more complex conditions and then using those conditions in the action preconditions. In either case the result is a more explicit description of the domain that is easier to understand, debug and modify.

### 2.2 (at end):

(at end) is a very strange condition. In fact, it did not appear in any of the competition domains. I would argue that it also is not needed; not because it is easily captured by other constructs, but rather because it is unnatural and would never appear in a reasonable domain model. (at end) is intended to support the flexibility whereby an action can "release" a condition so that other actions might delete that condition, as long as the condition is subsequently restored on time. Like the (over all) condition it has the effect of breaking the Markovian nature of normal action specifications. A more natural way to model such a situation, I would claim, is simply to use conditional (at end) effects: if the required condition holds at the end, the desired effect will be created, otherwise some bad effect will occur. The so modified load-truck action is given in Figure 2.

```
(:durative-action burnMatch
   :parameters (?m - match ?l - location)
   :duration (= ?duration 5)
   :precondition (and (have ?m)
                      (at ?l))
   :effect     (and (when (no-other-light-source ?l)
                       (and (not (dark ?l)) (light ?l)))
                    (not (have ?m))
                    (burning ?m)
                    (at end (when (burning ?m) (not (burning ?m))))
                    (at end (when (and (no-other-light-source ?l)
                                       (burning ?m))
                                  (and (not (light ?l))
                                       (dark ?l))))))
)
(:action blowOutMatch
   :parameters (?m - match ?l - location)
   :precondition (and (at ?l)
                      (burning ?m))
   :effect     (and (not (burning ?m))
                    (when (no-other-light-source ?l)
                       (and (not (light ?l))
                            (dark ?l))))
)
```

Figure 3: Alternate model of `burn-match`

In this modification, instead of a `(at end (holding ?c ?o))` condition, we have simply changed the effects of the action. If `(holding ?c ?o)` holds at the end of the action, the action has its normal effects. Otherwise, it adds to the state a marker indicating that a load has failed. If we add `(not (load-failed))` to the goal, then the planner would search for ways of falsifying the antecedent of the effect (assuming that `(load-failed)` cannot be undone), i.e., the planner would search for ways of ensuring that `(holding ?c ?o)` is true at the end of the action. Note that this is exactly what the planner would do to ensure that an ordinary precondition holds. That is, again we can reduce the construct down to standard features.

## 3. Unspecified Durations

Another feature of PDDL2.1 is the ability to specify ranges of durations for actions. The intent here is that the actual duration of the action might be affected by other actions. The `burnMatch` and `heat-water` actions (Figures 10 and 12 of the PDDL2.1 document) are examples where a range is utilized as the duration.

I find flexible durations to be strange, they again make the action dependent on future actions. Furthermore, I am not convinced that they are necessary. Rather I think that a more natural way to model such situations would be to introduce two actions, one to start the action (light the match, or start heating the liquid), and one to end the action (blow out the match, or take the liquid off the heat). Figures 3 and 4 present these alternate models.

The `burnMatch` starts the match burning (toggling the lighting status of the location if there was no other light at that location). It also posts a "default" completion of the

```
(:durative-action heat-water
   :parameters (?p - pan)
   :duration (= ?duration (/ (- 100 (temperature ?p)) (heat-rate)))
   :precondition (and (full ?p)
                      (onHeatSource ?p)
                      (byPan))
   :effect (and (heating ?p)
                (heating-start ?p (current-time))
                (at end (when (and (byPan) (heating ?p)) (not (heating ?p))))
                (at end (when (heating ?p)) (assign (temperature ?p) 100))
                (at end (when (and (not (byPan)) (heating ?p)) (burn-pot ?p)))))

(:action take-off-heat
   :parameters (?p - pan)
              (?startt - time)
   :precondition (and (heating ?p)
                      (heating-start ?p ?startt)
                      (byPan))

   :effect (and (not (heating ?p))
                (when (not (burn-pot ?p))
                    (increase (temperature ?p)
                              (* (- (current-time) ?startt) (heat-rate))))))
```

Figure 4: Alternate model of `heat-water`

match burning at the maximum duration. If the match is still burning at the end of this maximum duration it is extinguished and the lighting status is toggled (if the match was the only source of light). On the other hand, the match can be extinguished earlier by a `blowOutMatch` action.

The `heat-water` action starts the water heating, and like `burn-match` has a default maximum duration. If the pan is still being heated at the end of that time it raises the temperature of the pan to 100 degrees, and, if the agent is by the pan, it takes the pan off the heat, otherwise pan continues to be heated causing it to become burnt (this is in keeping with our previous discussion about not wanting (`at end`) preconditions). Note the pan's temperature never rises above 100 (we are assuming that the water keeps on boiling). It also marks the time that heating started ((`current-time`) is the time that the action was executed). `take-off-heat` is an action that can take the pan off the heat at any time. It uses the start time of the heating to calculate the temperature of the water (if the pot is burnt, its temperature remains at 100, as set by `heat-water`).

I am not suggesting that these alternate action specifications are the "right" models (e.g., the `heat-water` cannot account for putting the pot back on the heat after taking it off). What I am suggesting is that the case for variable durations has not been made. Is it really necessary or even natural for domain modeling?

## 4. Conclusions

I could list a few other components of pddl2.1 that seem to me to be unnecessary, but I believe that the point has been made. pddl2.1 is essential for the planning competition, and I am certainly a strong supporter of the usefulness of the competition for furthering planning research. However, I would suggest that outside of the context of the competition, the issue of which features should be included in a planning domain specification language needs to be more grounded in the application of these languages. Planning domains, even simplified ones designed for research, can be modeled in many different ways, and I believe that it is better to produce more robust models with simpler languages than to develop languages with features that are not really needed.

I do think that many of the ideas contained in PDDL2.1 are useful, e.g., the way in which continuous change is treated. Nevertheless, I would not encourage anyone to try to construct planning algorithms for dealing with these features. Rather I would encourage the development of planning algorithms inspired by the issues that arise from interesting domains. That is, I think that the incorporation of new features into planning languages needs to be motivated by compelling examples.